\documentclass{preprint}

\usepackage{array}      
\usepackage{xspace}
\usepackage{amsmath,amssymb,amsfonts}
\usepackage{booktabs}
\usepackage{graphicx}
\usepackage{tabularx}
\usepackage{xcolor}
\usepackage{tabularx}
\usepackage{spverbatim}
\usepackage{mdframed}
\usepackage{multirow}
\usepackage{colortbl}
\usepackage[numbers,comma,sort&compress]{natbib} 
\usepackage{hyperref}

\title{X-FEMR: A Token-level Explainable Approach for Electronic Health Records Foundation Models using Transformer-based Models}

\author[1,$\dagger$]{Jie Huang}
\author[1,$\dagger$]{Pengfei Yin}
\author[1]{Zihan Xu}
\author[1,2]{Daniel Capurro}
\author[1]{Mike Conway}
\author[1,*]{Ting Dang}
\affil[1]{School of Computing and Information Systems, University of Melbourne, Melbourne, Australia}
\affil[2]{Department of General Medicine, Royal Melbourne Hospital
Melbourne, Australia}

\affil[$\dagger$]{These authors contributed equally to this work.}
\affil[*]{Corresponding author(s). Email(s): \url{ting.dang@unimelb.edu.au}}

\newcounter{prompt}

\begin{document}

\maketitle

\begin{abstract}
Foundation Models for Electronic Health Records (FEMRs) are pretrained on large-scale structured patient data, enabling them to convert longitudinal patient trajectories into generalizable representations for diverse clinical prediction tasks. Despite their effectiveness, FEMRs remain black-box models, raising concerns about bias, interpretability, and clinical trust. To address this, we propose the first token-level explainability approach for FEMRs. We train a Transformer-based surrogate model on input-output pairs from the FEMR across two prediction tasks, approximating its behavior while preserving temporal dynamics. We identify the most influential tokens, providing insights into how FEMRs leverage different aspects of patient history for predictions. To evaluate clinical relevance, we introduce a novel clinical alignment metric that quantifies the correspondence between the surrogate model’s key tokens and clinically validated features. Our results demonstrate that the surrogate closely approximates FEMR predictions and that token-level explanations align well with clinical knowledge, offering a practical framework for interpretable and trustworthy clinical AI.

\end{abstract}


\section{Introduction}

Foundation models (FMs) are trained on large-scale and diverse datasets using self-supervised objectives, learning to acquire general-purpose  representations that can be efficiently adapted to a wide range of downstream tasks~\cite{Guo2025-en}. The use of abundant data, transformer-based architectures, and advances in optimization and model scaling has produced models with strong transferability and emergent capabilities~\cite{kaplan_scaling_2020}. Consequently, FMs often outperform task-specific systems, reducing the need for extensive task-dependent supervision and making them well-suited for complex, data-rich domains, such as healthcare. 

In the field of medical artificial intelligence (AI), FMs have primarily been developed along two major lines, reflecting the heterogeneous nature of clinical data. The first stream focuses on unstructured clinical text, creating Clinical Language Models (CLaMs), which are trained on narrative notes and reports. While CLaMs have demonstrated strong performance in extracting information from unstructured clinical narratives, a substantial portion of clinically relevant information is encoded in structured electronic health records (EHRs), which capture longitudinal diagnoses, procedures, medications, and their temporal relationships. Consequently, increasing attention has shifted toward FMs for structured EHR data. These models learn from longitudinal sequences of coded clinical events to capture temporal patterns in patient trajectories, which are referred to as FMs for EHRs (FEMRs)~\cite{wornow2023shaky}, include ETHOS~\cite{Renc2025-tr} and CLMBR-T-Base~\cite{wornow2023ehrshot}.

Despite their strong empirical performance, the translation of FEMRs into clinical practice remains limited. First, their internal representations and decision mechanisms are largely opaque~\cite{Wang2025-ec}, limiting interpretability for clinicians and model developers. Second, while interpretability tools for clinical language models have rapidly matured, they mainly focus on traditional machine learning approaches~\cite{SBAND2023101286,healthcare13172154}, and thus comparable explainability frameworks for FEMRs are still underdeveloped. Third, FEMRs are typically trained on institution-specific datasets constrained by privacy and governance policies, which can encode local practice patterns and biases, thereby raising concerns regarding robustness and generalizability across healthcare settings. Consequently, there is limited understanding of how these models reason over patient histories and temporal event sequences. Together, these limitations motivate the need for explainable FEMRs that provide transparent and clinically meaningful explanations of their representations and predictions. Evaluations for explainability in FEMRs are also lacking due to the absence of explainable methods for FEMRs.

A broad range of explainable AI (XAI) techniques have been proposed to interpret complex machine learning models. Prominent examples include feature-attribution methods such as SHAP~\cite{NIPS2017_8a20a862}, which quantify the contribution of individual input features to model predictions, as well as gradient-based approaches like saliency maps~\cite{simonyan_deep_2014}, and perturbation-based methods like LIME~\cite{ribeiro_why_2016}. However, their effectiveness is substantially limited when applied to FMs. FMs operate on high-dimensional, distributed representations and often rely on complex internal interactions across long contexts, making it difficult to meaningfully attribute predictions to individual input features. Moreover, post hoc attribution methods typically assume a fixed input–output mapping, whereas FMs produce task-agnostic latent representations that are reused across multiple downstream tasks. As a result, existing XAI techniques often yield explanations that are unstable, difficult to interpret clinically, or misaligned with the reasoning processes of FMs. This highlights the need for explanation methods that are specifically designed for foundation-scale architectures.

Recent studies have begun to explore XAI techniques for FEMRs, but this line of work has mainly focused on Large Language Models (LLMs) for general purpose. Saraswat et al.~\cite{Saraswat2022-jf} proposed a four-axis taxonomy for XAI, encompassing data explainability, model explainability, post hoc explainability, and explanation assessment. Within this framework, recent XAI research has emphasized the role of proxy representations, interpretable surrogate models that approximate the behavior of black-box systems. As discussed by~\cite{Saraswat2022-jf}, proxy models served as intermediate, human-aligned representations that enabled inspection of how a deep model responds to clinically relevant features, even when its internal states are not directly accessible. By introducing an interpretable bottleneck or fitting transparent surrogates such as linear or additive models, proxy representations map complex temporal EHR inputs into a lower-dimensional semantic space that clinicians can reason about. This approach facilitates attribution analysis, detection of spurious feature dependencies, and assessment of whether model behavior aligns with established clinical reasoning. However, the current surrogate models, 
such as linear regression, 
and Decision tree (DT), generally assume that the data is smooth linear~\cite{sarker2021machine}, which is not true for EHRs. 

Despite advances in these surrogate models, most approaches still focus mainly on feature-level explanations. In contrast, FMs generate predictions one token at a time using an autoregressive learning process. This allows for token-level explainability, which can be especially useful in clinical settings. For example, knowing which specific words or values in a patient’s record influenced a model’s prediction can help clinicians better understand the reasoning behind AI recommendations and improve clinical scoring methods or practice guidelines~\cite{pmlr-v106-tonekaboni19a}. Some recent studies have explored token-level explanations in general LLMs using methods like Integrated Gradients (IG), a gradient-based technique that assigns importance scores to each token in a text~\cite{axiomatic2017,ALI2023101805}. However, IG can be fragile in identifying the right features and may give inconsistent explanations~\cite{ALI2023101805}. Despite its potential, token-level explainability remains rare in clinical applications, particularly for longitudinal EHRs.

Another challenge in explaining predictions from EHR data is their temporal nature. Beyond identifying which clinical variables are important, it is important to know when they influence the prediction. Token-level explainability offers this finer-grained insight, identifying the specific variables and time points that drive model outputs. However, common surrogate models, such as logistic regression, struggle to capture these temporal dynamics, limiting their effectiveness for interpretable predictions in longitudinal EHR data. Furthermore, the evaluation of explainability itself remains an open challenge, with few quantitative metrics available to assess whether explanations align with clinical knowledge or decision-making.

In this work, we propose the first token-level explainability approach for FEMRs. Specifically, we introduce a transformer-based surrogate model to approximate the input-output behavior of FEMRs, enabling the capture of temporal dynamics in EHRs for more reliable explanations. In addition, we develop a clinically validated evaluation metric that quantitatively measures the quality of the model explanations. Our experiments on two distinct clinical datasets and tasks show that the surrogate model closely approximates the behavior of FMs in EHRs while achieving comparable predictive performance. Moreover, token-level explainability demonstrates stronger alignment with clinical tasks with the ratio of 0.3, suggesting the effectiveness of our approach. This work lays the foundation for more interpretable and clinically actionable AI in longitudinal EHR analysis.

\section{Related Work}
FMs trained on structured EHR data, such as ETHOS~\cite{Renc2025-tr}, CLMBR-T-Base~\cite{wornow2023ehrshot}, and Med-BERT~\cite{rasmy_med-bert_2021}, have demonstrated strong performance on a range of clinical tasks. These models support applications including treatment recommendation and 30-day hospital readmission prediction, particularly after task-specific fine-tuning~\cite{Almeida2026,TIMILSINA2025109925}. 

Traditional interpretable models, including probabilistic approaches such as Naïve Bayes and sequence models like Hidden Markov Models, provide intuitive explanations through explicit probability distributions and state transition structures. However, these models lack the capacity to capture the complexity of modern learning systems. Recent advances in explainable artificial intelligence have enabled post hoc interpretation of deep learning models, improving transparency and supporting clinical adoption~\cite{bienefeld_solving_2023}. Common approaches include feature attribution methods, gradient based explanations, and surrogate models. Nonetheless, these techniques remain limited when applied to foundation scale models.

Research on the interpretability of FMs is still sparse and has largely focused on medical imaging. More specifically, prior work has explored explainability for vision FMs in retinal and breast imaging applications~\cite{lee_optimizing_2025,cavalcante2025explainableaiapproachesbreast}. However, explainability methods for FEMRs are largely unexplored, leaving their internal reasoning processes insufficiently understood.

\section{Methodology}
In this section, we present our explainability pipeline for analyzing FEMRs. We first describe its overall design, highlighting the integration of proxy representations with token-level interpretation. We then apply the pipeline to FEMRs on clinically relevant prediction tasks to assess their explainability.

\subsection{Overview of Explainability Pipeline}
\begin{figure}[t]
  \includegraphics[width=\columnwidth]{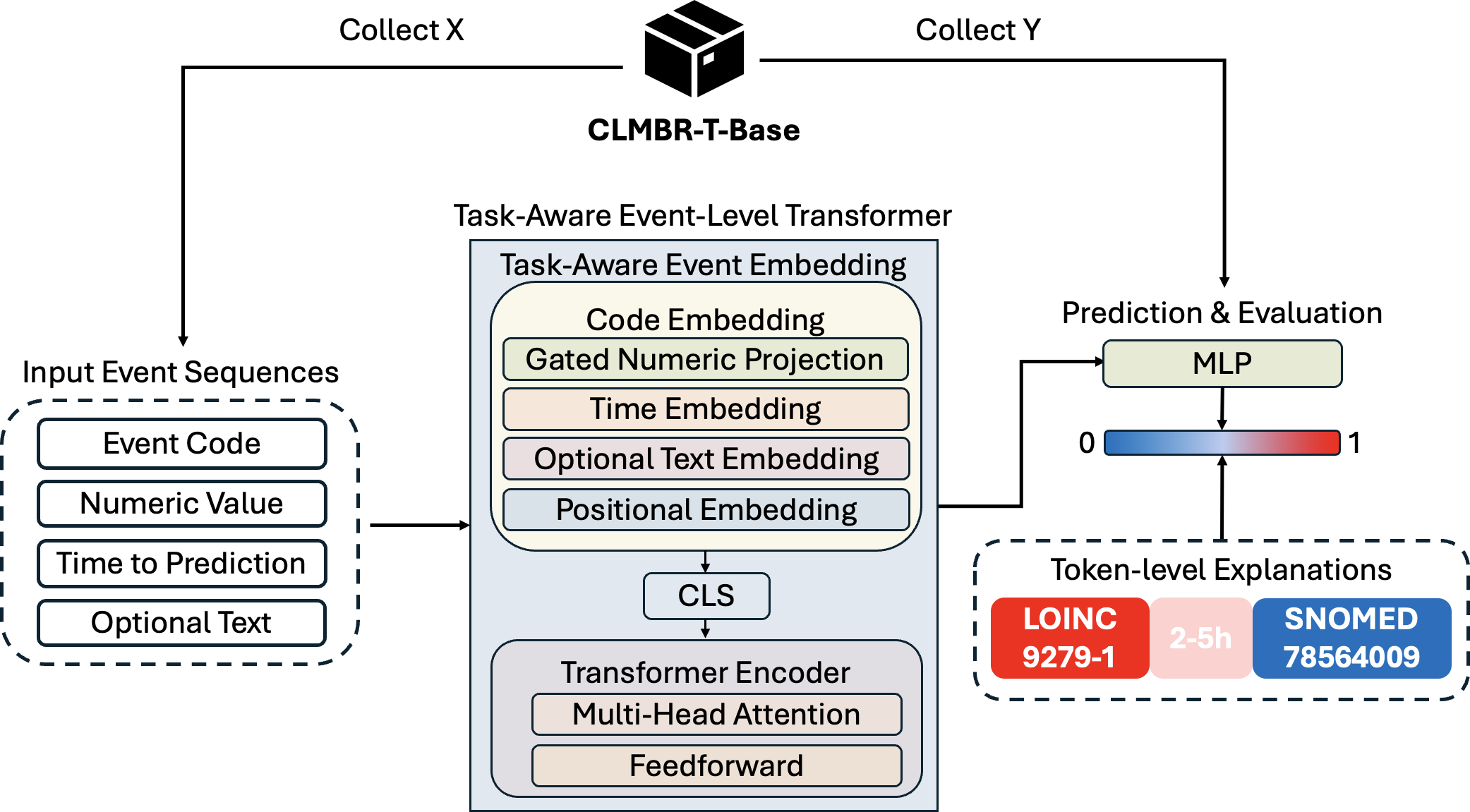}
  \caption{Pipeline of the study: We train a surrogate transformer using collected input and output pairs from the CLMBR-T-Base and then visualize the importance of tokens using SHAP analysis to implicitly understand how the FEMR trained on longitudinal data makes decisions in clinical prediction tasks. Based on the importance of tokens, we propose a new assessment method called the clinical validated events ratio to further measure the reliability of FM in clinical practice. 
  }
  \label{fig:overview}
\end{figure}

We propose an explainability pipeline that leverages a surrogate model to approximate the behavior of FEMRs and to enable explanations, as illustrated in Figure~\ref{fig:overview}. The pipeline trains a transformer-based surrogate on the input-output mappings of the target FEMRs and applies token attribution methods, such as SHAP, to quantify the contribution of individual token. This approach aims to open the black box of the FMs, revealing how it processes complex patient data to generate predictions. Our focus is primarily on token-wise explainability, assessing which clinical inputs most influence model outputs. This information helps clinicians interpret predictions and understand the factors driving model decisions. Additionally, we propose a novel clinical alignment metric that evaluates how closely the tokens identified by our surrogate models correspond to clinically validated features, thereby assessing the extent to which the surrogate’s explanations align with established medical knowledge.

To gain deeper insight into the FEMRs, we run the pipeline under two supervision settings. In the first setting, the surrogate is trained on the binary predictions of FEMRs as hard labels, capturing the model’s actual decisions. In the second, it is trained on the predicted probabilities as soft labels, reflecting the nuanced confidence of the models. Comparing token contributions and training dynamics across these settings allows us to distinguish the factors driving observed decisions from those influencing internal uncertainty, revealing where and how the model may deviate from optimal or clinically meaningful behavior. 
We will first present the data collection pipeline for the FMs, followed by the surrogate model development and explainability analyses.

\subsection{Data Collection from FEMRs}
\paragraph{Data format. }
FEMRs are trained on longitudinal, patient-level event sequences.
Each patient’s EHR can be viewed as a time-ordered sequence of clinical events, characterized by irregular time intervals, heterogeneous event types, and codes from multiple medical coding systems.
Some events additionally contain numeric values, such as laboratory measurements or vital signs, making EHR data both temporally and semantically complex.
In practice, although EHR data are sequential, they are typically stored in structured tables.
A widely adopted and standardized representation is the OMOP Common Data Model \footnote{https://www.ohdsi.org/data-standardization/} (OMOP-CDM), which organizes clinical events into relational tables with consistent formats across institutions.
In this work, we construct patient-level event sequences from raw data following the OMOP-CDM schema.
Each patient's trajectory is represented as chronologically ordered sequences of clinical events. Each event is characterized with four features:
\begin{itemize} 
    \item \textit{\textbf{Code ($c_t$)}} — e.g., SNOMED CT, LOINC, RxNorm  
    \item \textit{\textbf{Numeric value ($n_t$)}} — e.g., lab results  
    \item \textit{\textbf{Text field ($u_t$)}} — e.g., descriptions, notes, or categories  
    \item \textit{\textbf{Timestamp ($t_{event}$)}} — the time of the recorded event  
\end{itemize}

The code refers to standardized medical terminologies and coding systems used to describe and categorize various types of health information, such as LOINC for laboratory tests\cite{mcdonald_loinc_2003} and RxNorm for medications\cite{rxnorm}. The numeric and text fields contain additional information about the corresponding medical code. 

\paragraph{Collection of Input-Output Pairs. } 
For each prediction task, we define a prediction time $T_i$ for patient record $i$.
The input to the FEMR consists of the patient’s historical EHR data recorded \emph{prior to} $T_i$.
Formally, each patient record is represented by a time-ordered sequence of clinical events as: 

\begin{equation}
\mathcal{X}_i = \{ e_1, e_2, \dots, e_{t_k} \}, \quad \text{where } t_k < T_i
\end{equation}
and each event $e_{t_k} = (c_k, n_k, u_k, t_k)$ includes a medical code~$c_k$, a numeric value~$n_k$, a text field~$u_t$, and a timestamp~$t_{event}$. The FEMR encodes the event sequence $\mathcal{X}_i$ into a patient-level representation and produces a prediction for the target task.
This process produces two forms of output, which are continuous probabilities (soft labels) and binary labels based on a 0.5 threshold (hard labels) for classification tasks. Both types of labels are used with their corresponding inputs as pairs to train the surrogate model and to evaluate token alignment and interpretability.

\subsection{Surrogate Model Development}
EHR data is typically long and contains irregular, heterogeneous variables and nonlinear patterns, which simple surrogate models cannot capture effectively. Therefore, we employ a transformer-based surrogate model, which can represent these complex sequences and construct accurate proxy representations of the FEMR.

We propose an event-level transformer encoder to model longitudinal EHR sequences. Each patient record is represented as an ordered sequence of clinical events occurring before the prediction time, and the model produces a sequence-level prediction. 

To capture the irregular time gaps between events, we represent each event by its time-to-prediction interval, enabling the model to reason about the relative timing of events. 
Specifically, given the event timestamp $t_{event}$ and the prediction time $T$, we define 
\begin{equation}
\Delta t = \frac{T - t_{\text{event}}}{3600}
\end{equation}
which is measured in hours.
To mitigate the long-tailed distribution of time intervals, $\Delta t$ is log-transformed and normalized using statistics estimated from the training set, and used as a continuous time feature.

Then, each raw clinical event contains four different feature types (code, numeric\_value, text\_value, time\_delta) shown as:
\begin{equation}
x_t = (c_t, n_t, u_t,\Delta t)
\end{equation}

For encoding part, each event token is initialized with a learned code embedding, augmented with a text embedding.
Numerical values are included via a gated injection mechanism, which means the projected numeric embedding is adjusted by a content-dependent gate computed from the current event representation.
To encode temporal information, we project the normalised time-to-prediction feature of each event into the model dimension and adding it to the token representation.
Learnable positional embeddings are added, followed by layer normalization and dropout. 
\begin{equation}
\begin{aligned}
    \mathbf{e}_t =\;&
    \mathbf{E}_{\text{code}}(c_t)
    + \mathbf{E}_{\text{text}}(u_t)
    + \mathbf{g}(\cdot)\odot W_{\text{num}}\, n_t\\
    &+ W_{\Delta t}\, \Delta t_t 
    + \mathbf{E}_{\text{pos}}(t)
\end{aligned}
\end{equation}
The sequence of event embeddings is processed by a multi-layer transformer encoder to capture temporal dependencies among events.
A [CLS] token is then prepended to summarize the sequence, and the final [CLS] representation is fed into a task-specific prediction head for hard-label or soft-label supervision.

\paragraph{Hard-label Supervision.}
In this setting, the surrogate model is trained to predict the binary outputs of FEMR (e.g., ICU transfer vs. non-transfer). This approach focuses on capturing the final decision boundary of FEMR.

\paragraph{Soft-label Supervision.} 
In this approach, the surrogate model is trained to predict the raw probability outputs of the FEMR, enabling a more nuanced approximation of the  confidence and internal decision-making process of the FMs. This extends beyond binary classification to suit multiclass classification and regression tasks. 

This comparison further allows us to analyze how the choice of supervision influences the learned token importance patterns, providing insights into how closely the surrogate reflects the underlying decision logic of FMs.
We use both surrogate models for subsequent SHAP-based token attribution analysis.

\subsection{Token-Level Clinical Explainability}

To evaluate how well the surrogate model explains the CLMBR-T-Base in a clinically meaningful way, we perform a token-level analysis using SHAP. The goal is to determine whether the token identified as important by the surrogate model correspond to known clinical predictors of downstream tasks.

For each patient and each event $i$ in the EHR sequence $j$, SHAP assigns a value to each token that quantifies its contribution to the model prediction. This produces a map of token importance across all events in the dataset.
Meanwhile, we define a set of clinically relevant tokens based on different clinical evidence. 

To quantify the alignment between SHAP attributions and clinical knowledge, we count how often each token is assigned importance by the surrogate model: 
\begin{equation}
 c(f_j; D) = \sum_{x_i \in D} \mathrm{occurrence}(f_j, x_i),
\end{equation}
where \(D\) is the dataset and \(\mathrm{occurrence}(f_j,x_i)\) is the number of times token \(f_j\) appears in patient record \(x_i\).  
The clinical validated events ratio is then defined as:  
\begin{equation}
R = \frac{\sum_{f_j \in \mathcal{F}_{\mathrm{c}}} c(f_j; D)}{\sum_{f_j \in \mathcal{F}} c(f_j; D)}
\end{equation}
\noindent where \(\mathcal{F}\) is the set of all tokens from SHAP analysis, and \(\mathcal{F}_{\mathrm{c}} \subseteq \mathcal{F}\) denotes the subset of clinically relevant tokens. A higher \(R\) indicates greater alignment of the surrogate model’s explanations with established clinical knowledge.

We compute \(R\) for surrogate models trained under hard-label (binary predictions) and soft-label (predicted probabilities) supervision. This allows us to evaluate which surrogate better captures clinically meaningful reasoning from the CLMBR-T-Base. This analysis converts abstract SHAP values into a quantitative, interpretable metric. It enables us to assess not only whether the surrogate reproduces the predictions of CLMBR-T-Base, but also whether the tokens driving these predictions are consistent with known clinical knowledge.

\section{Experimental Setup}

\subsection{Patient Cohort and Tasks}
\paragraph{Cohort.} We used the EHRSHOT dataset due to its large-scale, longitudinal EHR records, which provide rich temporal information for modeling patient trajectories~\cite{wornow2023ehrshot}. The dataset contains records for 6,739 patients with a wide range of trajectory lengths. The longest patient sequence includes 199,913 events, reflecting the high variability and complexity of real-world EHR data. To ensure compatibility with transformer-based surrogate models to handle reasonable long sequences, 
we selected the top 75\% of patient records based on trajectory length. 

\paragraph{Tasks.} We focus on two classification tasks: long length of stay(LOS) and ICU transfer. The first task including predicting whether a patient’s total LOS during a visit to the hospital will be at least 7 days. LOS is a key factor of both patient outcomes and healthcare costs~\cite{measuring2025}. The prediction time is at 11:59pm on the day of admission, and visits that last less than one day (i.e. discharge occurs on the same day of admission) are excluded. The second task involves predicting whether a patient will be transferred to the ICU during their hospital stay. Predictions are made at 11:59 pm on the day of admission, and ICU transfers occurring on the admission day are excluded~\cite{wornow2023ehrshot}. Both tasks represent common clinical prediction problems that focus on forecasting patient outcomes. 

The final study cohorts after preprocessing consist of 4,868 and 5,275 longitudinal records for LOS and ICU transfer prediction tasks, respectively.
As shown in Table~\ref{tab:trajectory_length}, the length of patient trajectories varies substantially, ranging from 6 to 7,888 clinical events, with an average of $\sim$2,400 events.

~

\begin{table}[t]
\centering
\small

\begin{tabular}{lcc}
\toprule
\textbf{Statistic} & \textbf{LOS} & \textbf{ICU Transfer} \\
\midrule
Min   & 6    & 6    \\
Mean  & 2,453 & 2,418 \\
75\%  & 3,928 & 3,838 \\
Max   & 7,888 & 7,888 \\
\bottomrule
\end{tabular}
\caption{Distribution of patient trajectory lengths (measured as the number of events per patient) for two tasks.}
\label{tab:trajectory_length}
\end{table}

\subsection{Model Implementation}

\paragraph{Foundation Model.} 
The FEMR used for our explainability analysis is CLMBR-T-Base~\cite{guo2024multi}, a 141M-parameter autoregressive model trained on 2.57 million de-identified longitudinal EHRs from the EHRSHOT dataset~\cite{wornow2023ehrshot} at Stanford Medicine. CLMBR-T-Base is currently the only FEMR with publicly available weights, enabling reproducible analyses. 

\paragraph{Surrogate Model.} 
For both surrogate modeling tasks, we split the dataset into approximately 80\% training, 10\% validation, and 10\% test sets, using stratified sampling to preserve the label distribution across splits.
The split is performed at the patient level to ensure that records from the same patient do not appear in multiple sets, thereby preventing potential data leakage.
For modeling, we use a Transformer encoder with 3 layers, and each layer has a hidden dimension of 128 and 4 attention heads.
The feedforward network size is 256, and GELU (Gaussian Error Linear Unit) is used as the activation function.
A dropout rate of 0.2 is applied, and pre-normalization is adopted to improve training stability.
The maximum input sequence length is set to 2048.
Under both training strategies, the model is trained using the Adam optimizer with a learning rate of $10^{-4}$, a batch size of 16, and up to 100 training epochs.
Early stopping is applied based on the validation Area Under the Precision–Recall Curve (AUPRC), and training is terminated if no improvement is observed for 20 consecutive epochs.
The model is trained on a single GPU, which accelerates training for long sequences and Transformer-based models.

To further improve surrogate model training with soft labels, we apply a logit transformation to the predicted probabilities produced by the CLMBR-T-Base. This transformation maps probability values into the log-odds space, expanding small differences that would otherwise be compressed near zero or one. By operating in the logit domain, the soft labels provide richer supervisory signals and yield more informative gradients during backpropagation, leading to more stable and effective surrogate model optimization.

\subsection{Evaluation}
\begin{table}[t]
\small
\centering

\begin{tabular}{p{7cm}p{5cm}}
\hline
\toprule

\textbf{Clinical Validated Features} & \textbf{Code} \\ 
\toprule
Patient's diagnosis, comorbidities & ICD-10-CM\footnotemark[1]\\ \hline
Hospital Frailty Risk Score & 109 specific ICD-10 diagnostic codes found in a patient's hospital records (current and past 2 years) \\ \hline

Physiological measurement (including heart rate, blood pressure, temperture, oxygen saturation)
&
National Early Warning Score (including\newline 
LOINC/9279-1 
LP21258-6 \newline
LOINC/88658-0\newline
LOINC/LP201613-9\newline
LOINC/LP409229-4\newline
LOINC/LA17976-4\newline
LOINC/LA19825-1 \newline
LOINC/2019-8\newline
SNOMED/413444003 \newline
LOINC/8310-5\newline
LOINC/8480-6\newline
SNOMED/271649006 \newline
LOINC/8867-4\newline
SNOMED/78564009 \newline
LOINC/38214-3) \\ 
\bottomrule
\end{tabular}

\caption{\textbf{LOS clinical validated features and corresponding codes}~\cite{DESCHEPPER2025105678}. This table shows the clinically validated features used for LOS prediction tasks in explainability evaluations and corresponding codes. This provides clinical evidence to evaluate explainability using the proposed clinical alignment metric.}
\label{tab:clinical concept_los}
\end{table}
\footnotetext[1]{ICD-10-CM: International Classification of Diseases, 10th Edition, Clinical Modification}

\begin{table}[t]
\small

\centering
\begin{tabular}{p{4cm}p{3cm}}
\hline
\toprule
\textbf{Clinical Validated Features} & \textbf{Code} \\ 
\toprule
Respiratory rate & LOINC/9279-1 \\ \hline
Oxygen saturation & LP21258-6 \\ \hline
Use of supplemental oxygen &
LOINC/88658-0\newline
LOINC/LP201613-9\newline
LOINC/LP409229-4\newline
LOINC/LA17976-4\newline
LOINC/LA19825-1 \\ \hline
Hypercapnic respiratory failure &
LOINC/2019-8\newline
SNOMED/413444003 \\ \hline
Temperature & LOINC/8310-5 \\ \hline
Systolic blood pressure &
LOINC/8480-6\newline
SNOMED/271649006 \\ \hline
Heart rate &
LOINC/8867-4\newline
SNOMED/78564009 \\ \hline
AVPU (Alert, Voice, Pain, Unresponsive) Score &
LOINC/38214-3 \\
\bottomrule
\end{tabular}

\caption{\textbf{ICU transfer clinical validated features and corresponding codes}~\cite{news2012,smith_national_2019,verma_toward_2024}. This table shows the clinically validated features used for ICU transfer prediction tasks in explainability evaluations and corresponding codes.}
\label{tab:clinical concept_icu}
\end{table}

\paragraph{Clinical Validated Features.}

Multiple studies have proven that patient's diagnosis, laboratory, comorbidity variables and severity of illness(SOI) are all strong predictors of LOS~\cite{DESCHEPPER2025105678,street_association_2025,gokhale_hospital_2023,liu_comparison_2019}. We summarized those validated clinical variables in Table ~\ref{tab:clinical concept_los}~\cite{DESCHEPPER2025105678}. This provides clinical evidence to evaluate explainability using the proposed clinical alignment metric. 

For ICU transfer, it is a significant predictor of mortality, morbidity, and healthcare costs~\cite{young_inpatient_2003}. 
The task about ICU transfer are related to the National Early Warning Score (NEWS2), which aggregates vital signs and physiological measurements known to predict early clinical deterioration, including six physiological parameters: respiratory rate, oxygen saturation, supplemental oxygen, hypercapnic respiratory failure, temperature, systolic blood pressure, heart rate, and AVPU score~\cite{news2012,smith_national_2019,verma_toward_2024}. Table~\ref{tab:clinical concept_icu} displaying the clinically validated features used for ICU transfer tasks in explainability evaluations~\cite{hripcsak_george_observational_2015}.
~\begin{table*}[t]
\centering

\begin{tabular}{lcccccc}
\toprule
\textbf{Prediction Task} & \textbf{AUROC} & \textbf{AUPRC} & \textbf{Accuracy} & \textbf{Precision} & \textbf{Recall} & \textbf{F1-score} \\
\midrule
LOS
& 0.7046 & 0.4173 & 0.7428 & 0.4556 & 0.3140 & 0.3717 \\
ICU Transfer
& 0.7282 & 0.1592 & 0.9525 & 0.3125 & 0.0704 & 0.1149 \\
\bottomrule
\end{tabular}
\caption{Prediction performance of the CLMBR-T-Base model on two clinical prediction tasks.}
\label{tab:fm_results}
\end{table*}

\section{Results}
\subsection{CLMBR-T-Base Prediction Results}

The CLMBR-T-Base predictive ability on two tasks is summarized in Table ~\ref{tab:fm_results}. 
For the LOS dataset, it contains 1,274 positive samples out of 5,257, giving a positive rate of 24.2\%.
The CLMBR-T-Base achieves an AUROC of 0.7046 and an AUPRC of 0.4173, and shows a precision of 0.4556 and a recall of 0.3140.
Overall, the CLMBR-T-Base demonstrates a reasonable performance on the LOS prediction task as it achieves an AUPRC nearly two times higher than the baseline positive rate (24.2\%), suggesting meaningful performance gains. 
To provide further context for these results, previous studies on predicting prolonged length of stay have reported AUROC in the range of 0.72-0.77 and AUPRC around 0.34-0.36 for a wide range of machine learning and deep learning models on EHR datasets ~\cite{ruan2025towards}. 
Although the datasets differ, this suggests that CLMBR-T-Base performs competitively against existing benchmarks in the literature.

The ICU transfer dataset is highly imbalanced, with only 213 positive samples out of 4,868, giving a positive rate of only 4.4\%. 
It can be seen from the Table~\ref{tab:fm_results}, the CLMBR-T-Base achieves an AUROC of 0.7282 and an AUPRC of 0.1592.
Although the overall classification performance remains challenging, as indicated by the low recall (0.0704) and F1 score (0.1149), the model overperforms the baseline positive rate.
It achieves an AUPRC that is more than three times higher than the prevalence of ICU transfer events.
Previous studies on ICU transfer tasks have often reported the AUROC as the primary evaluation metric, with values ranging from 0.85-0.90 ~\cite{wellner2017predicting}.
Although the AUROC achieved by CLMBR-T-Base on this task is lower than that reported for some other models, direct comparison is difficult due to differences in datasets used and the definitions of the outcomes.
Moreover, AUROC alone may be inadequate for understanding model performance in highly imbalanced settings.
In this context, the substantial improvement in AUPRC suggests that, despite the severe class imbalance, CLMBR-T-Base can capture informative signals for ICU transfer risk.

~


\begin{table*}[t]
\centering
\setlength{\tabcolsep}{16pt} 

\begin{tabular}{lcccc}
\toprule
& \multicolumn{2}{c}{\textbf{LOS}} & \multicolumn{2}{c}{\textbf{ICU Transfer}} \\
\cmidrule(lr){2-3} \cmidrule(lr){4-5}
\textbf{Metric} 
& \textbf{Hard-label} & \textbf{Soft-label} 
& \textbf{Hard-label} & \textbf{Soft-label} \\
\midrule
AUROC     
& \textbf{0.7637} & \textbf{0.7968} 
& \textbf{0.8235} & \textbf{0.9351} \\
AUPRC     
& \textbf{0.4434} & 0.3903 
& \textbf{0.2701} & \textbf{0.2656} \\
Precision 
& 0.4104 & 0.3429 
& \textbf{1.0000} & 0.0000 \\
Recall    
& \textbf{0.6044} & 0.2105 
& \textbf{0.2500} & 0.0000 \\
Accuracy  
& \textbf{0.7677} & \textbf{0.8626} 
& \textbf{0.9938} & \textbf{0.9959} \\
F1-score  
& \textbf{0.4889} & 0.2609 
& \textbf{0.4000} & 0.0000 \\
\bottomrule
\end{tabular}
\caption{Prediction performance of the best-performing surrogate model under hard-label and soft-label supervision for two prediction tasks.}
\label{tab:surrogate_both_tasks}
\end{table*}

\subsection{Surrogate Model Comparison}
To validate whether the support model can match or exceed the performance of the FEMRs and ensure reliable explainability, we also evaluated the performance of the transformer models on both tasks, as shown in Table~\ref{tab:surrogate_both_tasks}, using both hard labels and soft labels. For the LOS task, the performance on hard labels shows better results than the FEMR, demonstrating the effectiveness of the surrogate model. Specifically, the surrogate model's AUPRC achieves 0.4434 compared to CLMBR-T-Base's 0.4173, and the surrogate model's F1-score achieves 0.4889 compared to CLMBR-T-Base's 0.3717, indicating that the surrogate model can provide reasonable and reliable approximation of FM's predictive behavior. Regarding the soft labels, it underperforms compared to the CLMBR-T-Base, showing the surrogate model's AUPRC achieves 0.3903 compared to CLMBR-T-Base's 0.4173. This is possibly due to soft-label supervision relies on probability outputs, where predictions slightly above decision threshold (e.g., 0.5) are actually weak positives. A similar trend is observed for the ICU transfer tasks. Therefore, we adopted the surrogate model under hard-label supervision for the following explainability analysis.

\subsection{SHAP Analysis Results}

\begin{table}[t]
\centering
\small

\begin{tabularx}{\columnwidth}{@{}l X r@{}}
\toprule
\textbf{Prediction Task} & \textbf{Code (Event Name)} & \textbf{Frequency} \\
\midrule

\multirow{5}{*}{LOS} 
  & LOINC/8867-4(Heart rate)& 806 \\
  & LOINC/8480-6(SBP) & 412 \\
  & LOINC/8310-5\newline
  (Body temperature)& 325 \\
  & LOINC/8462-4(DBP)& 308 \\
  & LOINC/9279-1(Respiratory rate)& 258 \\
\midrule

\multirow{5}{*}{ICU transfer}
  & LOINC/8867-4(Heart rate)& 581 \\
  & LOINC/8310-5\newline
  (Body temperature)& 405 \\
  & LOINC/8480-6(SBP)& 314 \\
  & LOINC/8462-4(DBP)& 272 \\
  & LOINC/LP21258-6\newline
  (Oxygen saturation)& 214 \\
\bottomrule
\end{tabularx}
\caption{Top 5 most frequent codes in tokens}
\label{tab:top5_single}
\end{table}

\begin{table}[t]
\centering
\small

\begin{tabularx}{\columnwidth}{X c c}
\toprule
\textbf{Prediction Task}&
\textbf{No. of Clinical Validated Events} & \textbf{Ratio} \\
\midrule

LOS
& 2,257 & 0.3093 \\

ICU transfer 
& 1,985 & 0.2717 \\

\bottomrule
\end{tabularx}
\caption{Clinical validated events ratios associated with LOS and ICU transfer in surrogate models. Total events for LOS is 7,296 while total events for ICU transfer is 7,305.}
\label{tab:shap_combine}
\end{table}

For four features of each event, we find that numerical values (e.g., laboratory measurements) and time delta (i.e., duration of events) contributed most significantly, while text and code contributed less. This is because these two features contain the patient's clinical information: numerical values indicate whether the patient experienced the corresponding clinical event and the specific value of the event; time differences represent the patient's medical history over time.

\paragraph{Clinical Validated Events Ratio}

We summarize the most common codes in tokens in Table~\ref{tab:top5_single} and list the clinical validated events ratio for these two prediction tasks in Table~\ref{tab:shap_combine}.

Table~\ref{tab:top5_single} shows that the most influential tokens are primarily composed of clinical events; the top five are separately heart rate, systolic blood pressure (SBP), body temperature, diastolic blood pressure (DBP), respiratory rate, and oxygen saturation. We find that the most frequent codes in tokens correspond to clinical validated features we investigated in two prediction tasks, including heart rate, blood pressure, respiratory status, oxygen saturation. This provides token-level evidence that the model's predictions are primarily driven by these clinical validated events, and our identified tokens are clinically meaningful. 

Table~\ref{tab:shap_combine} shows the clinical alignment metric associated with LOS and ICU transfer in surrogate models. In the LOS prediction task, the clinical validated events are 2,257, ratio is 0.3093. This number corresponding to the top 5 most frequent codes in tokens totaled 2,109, accounting for 29\% of the total events. A similar trend is observed in the ICU transfer prediction task, showing the top 5 most frequent codes in tokens totaled 1786, representing 24\% of the total events, while clinical validated events are 1,985, accounting for 27.17\%. That implies the clinical validated events are highly concentrated in the most frequent clinical codes.

\section{Discussion}

Our surrogate design allows us to examine not only what the FEMR predicts, but how it uses different tokens to make those decisions. More importantly, this comparison moves beyond traditional performance metrics and provides an interpretable, token-level evaluations of how the surrogate model aligns with the FEMR. This clinical validated events ratio quantifies an approximate estimate of how many validated clinical events are involved in the model prediction, which helps selecting clinically meaningful events for model training in the future. However, the limitation in this study lies in the alignment problem between the original black box model and the surrogate model. This mismatch can be attributed to several factors: (1) longitudinal patient trajectories are highly complex and noisy, which contains multiple data modalities and irregular temporal information, (2) the patient trajectories are long, which makes sequence-level imitation challenging, and (3) the transformer-based surrogate model typically requires large amount of training data, while the number of available longitudinal EHRs is limited.

Another limitation of the current framework is that clinical events are represented by combining multiple feature channels into a single event-level token.
Although effective, this design can result in the loss of fine-grained token-level information.
Also, token-level explanations may overemphasize individual events while overlooking clinically meaningful trends used by FEMRs to perform prediction.
Future work could explore alternative tokenization strategies that treat time as a primary modeling component.
For instance, temporal information could be represented as a distinct token type in order to explicitly model the irregular intervals between events, in a manner similar to recent transformer approaches based on trajectories ~\cite{Renc2024-ga}. 
In the future, as more FEMRs become publicly available, we will generalize our approach to other foundation models.

\section{Conclusion}

In this work, we present the first token-level explainability approach for FEMRs, providing insights into the specific tokens in EHR data that drive model predictions. We also introduce a novel clinical alignment metric to quantify how well the tokens identified by our surrogate model correspond to clinically validated features, evaluating the alignment of model explanations with established medical knowledge. Our approach addresses a critical gap in the literature on interpreting FEMRs and offers a foundation for future studies aimed at assessing the reliability and clinical trustworthiness of Foundation Models in healthcare.



\section*{Contribution Statement}
Jie Huang and Pengfei Yin are designated as co-first authors.


\bibliographystyle{named}
\bibliography{ijcai26}



\end{document}